\documentclass[10pt,twocolumn,letterpaper]{article}

\usepackage{iccv}
\usepackage{times}
\usepackage{epsfig}
\usepackage{graphicx}
\usepackage{amsmath}
\usepackage{amssymb}

\usepackage{subfig}
\usepackage{multirow}
\usepackage{subfig}
\usepackage{dsfont}
\usepackage{booktabs}
\usepackage[breaklinks=true,bookmarks=false]{hyperref}

\iccvfinalcopy 

\ificcvfinal\fi

\begin{document}

\title{Distilling Pixel-Wise Feature Similarities for Semantic Segmentation}

\author{Yuhu Shan\\
{\tt\small tiger.yuhu.shan@gmail.com}
}

\maketitle
\begin{abstract}
Among the neural network compression techniques, knowledge distillation is an effective one which forces a simpler student network to mimic the output of a larger teacher network. However, most of such model distillation methods focus on the image-level classification task. Directly adapting these methods to the task of semantic segmentation only brings marginal improvements. In this paper, we propose a simple, yet effective knowledge representation referred to as pixel-wise feature similarities (PFS) to tackle the challenging distillation problem of semantic segmentation. The developed PFS encodes spatial structural information for each pixel location of the high-level convolutional features, which helps guide the distillation process in an easier way. Furthermore, a novel weighted pixel-level soft prediction imitation approach is proposed to enable the student network to selectively mimic the teacher network's output, according to their pixel-wise knowledge-gaps. Extensive experiments are conducted on the challenging datasets of Pascal VOC 2012, ADE20K and Pascal Context. Our approach brings significant performance improvements compared to several strong baselines and achieves new state-of-the-art results. 
\end{abstract}

\section{Introduction}
\begin{figure}[t]
	\begin{center}
		\includegraphics[width=8cm, height=5cm]{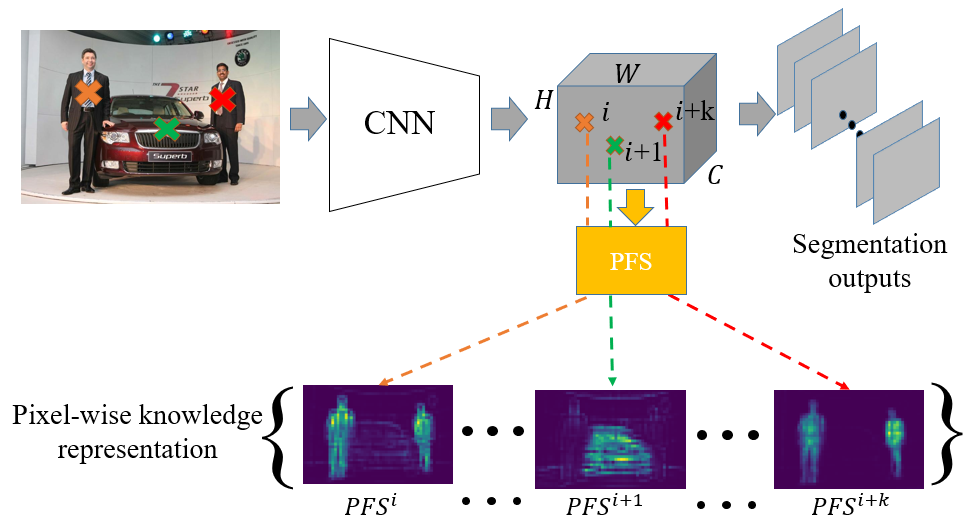}
	\end{center}
	\vspace{-0.5cm}
	\caption{Pixel-wise feature similarities for different high-level feature locations marked with different colored crosses. Brighter color in the map PFS$^{i}$ means stronger similarities between current feature location $i$ and all the other ones. The final distilled knowledge representation includes all the similarity maps with number of $H\cdot W.$}
	\label{fig1}
	\vspace{-0.5cm}
\end{figure}
Deep learning methods have been successfully applied to a wide range of tasks in computer vision, \emph{e.g.}, image classification~\cite{krizhevsky2012imagenet,he2016deep}, object detection \cite{ren2015faster} and semantic segmentation \cite{long2015fully}. The outstanding results are usually achieved by employing deep neural networks with large capacity. However, these large and deep neural networks \cite{he2016deep,szegedy2017inception} are difficult to be deployed in edge-devices with severe constraints on memory, computation and power, especially for the dense prediction tasks of semantic segmentation or scene parsing. Knowledge distillation technique is a popular and effective solution to improve the performance of a smaller student network by forcing it to mimic the behaviors of the original complicated teacher network, which is promising for offering deployable efficient network models.

Existing knowledge distillation methods are mostly for image classification where a student network tries to learn the knowledge represented by the teacher networks, such as soft predictions ~\cite{hinton2015distilling}, logits ~\cite{ba2014deep}, or features in intermediate layers ~\cite{romero2014fitnets,li2017mimicking,chen2017learning}. However, few works consider knowledge distillation for the dense pixel-level prediction tasks, \emph{e.g.}, semantic segmentation. Directly extending current distillation methods to this complicated task can only bring marginal improvements, since most of them provide image-level global knowledge, while pixel-wise knowledge is needed for semantic segmentation. 
Naively, intermediate convolutional neural network (CNN) features can be regarded as pixel-wise knowledge represented by the fixed feature vectors.
However, it is quite challenging for the student to learn the exactly same feature values across all of the pixel locations for semantic segmentation, due to the inherent capacity gap between the student and the teacher. Only marginal improvements can be obtained as shown by our experimental results. 
Therefore, new pixel-wise knowledge representation should be developed to ease the whole distillation process of semantic segmentation. 

In this paper, we propose to utilize pixel-wise feature similarities (PFS) between different pixel locations of the high-level CNN features as the knowledge transfer media for distillation. As shown in Figure \ref{fig1}, for each space location $i$ of the CNN feature maps, we endow it with a calculated PFS${^i}$ map. After visiting all the feature locations, we obtain the final generated knowledge representation, which includes all the PFS maps with the number of $H\cdot W$.  
Compared to the naive CNN features, our proposed PFS owns two advantages. The first one is that PFS models intra relationships of different image contents, which are invariant to the network architectures. In contrast, high-level CNN feature values of the teacher network are variant when using different network architectures. The second advantage is that PFS encodes more image contexts than the naive CNN feature vectors, which have been certified beneficial to the task of semantic segmentation\cite{zhao2017pyramid}.     
To further promote the distillation performance, we propose another new knowledge-gap guided knowledge distillation approach for the soft prediction mimicking, which also encourages more knowledge transfer in the pixels where large gaps exist between both networks. 

To summarize, our contributions include:

1. We propose a novel way of learning pixel-wise feature similarities for the knowledge distillation of semantic segmentation.     

2. A novel knowledge-gap based distillation approach on the soft predictions is proposed to achieve the weighted knowledge transfer for the pixels, based on the student's necessity of be taught by the teacher network at each pixel.

3. Extensive analysis and evaluations of the proposed approach are provided on the challenging ADE20K, Pascal VOC 2012 and Pascal Context benchmarks, to validate the superiority of the approach over the conventional knowledge distillation methods and current state-of-the-arts. Significant performance improvements are achieved by our newly proposed method.

\section{Related Work}
\subsection{Semantic segmentation}
Semantic segmentation, as a fundamental problem in computer vision, has been largely studied in the past several years. Currently, deep learning based methods occupy the best entries of several semantic segmentation benchmarks, \emph{e.g.}, Pascal VOC, ADE20K, and Pascal Context. \cite{long2015fully} is the pioneer work that address semantic segmentation with deep convolutional networks. It presents a conv-deconv architecture to achieve the end-to-end segmentation with fully convolutional networks. After this, many following works are proposed to further enhance the model's performance. For example, \cite{liu2015parsenet} tries to augment the features at each pixel location with the global context features. \cite{yu2017dilated,chen2018deeplab} introduce dilated operation into the convolutional operation to enlarge the filters' receptive field, which is testified beneficial for the segmentation task.  
\cite{zhao2017pyramid} proposes spatial pyramid pooling to aggregate the information from different global scales to help segmentation. \cite{chen2017rethinking} proposes DeepLab v3, a state-of-the-art segmentation model by introducing parallel dilated spatial pyramid pooling to the image-level features for capturing multi-scale information. Despite the fact that superior performance is achieved, most of the models require larger backbone networks or computationally expensive operations which is too heavy to be deployed in edge-devices. Therefore, many works are proposed recently to accelerate the networks for practical applications, which are reviewed in the following Section \ref{net_acc}.   
\subsection{Network acceleration} \label{net_acc}
Neural network acceleration approaches can be roughly categorized into weight pruning~\cite{han2015deep,szegedy2016rethinking}, weight decomposition~\cite{canziani2016analysis,novikov2015tensorizing}, weight quantization~\cite{gong2014compressing,rastegari2016xnor} and knowledge distillation~\cite{hinton2015distilling,romero2014fitnets,huang2017like}. Weight pruning methods~\cite{han2015deep,szegedy2016rethinking} iteratively prune the neurons or connections of low importance in a network based on designed criteria. Weight decomposition~\cite{canziani2016analysis,novikov2015tensorizing} converts the dense weight matrices to  a compact  format to reduce the parameter number greatly. Weight quantization~\cite{gong2014compressing,rastegari2016xnor} reduces the precision of the weights or features and then the generated low-precision model can be deployed during inference. However, most of weight quantization methods need specific hardware or implementation, and thus cannot fully capitalize modern GPU and existing deep learning frameworks.
\begin{figure*}[t]
	\begin{center}
		\includegraphics[width=17.5cm, height=8.5cm]{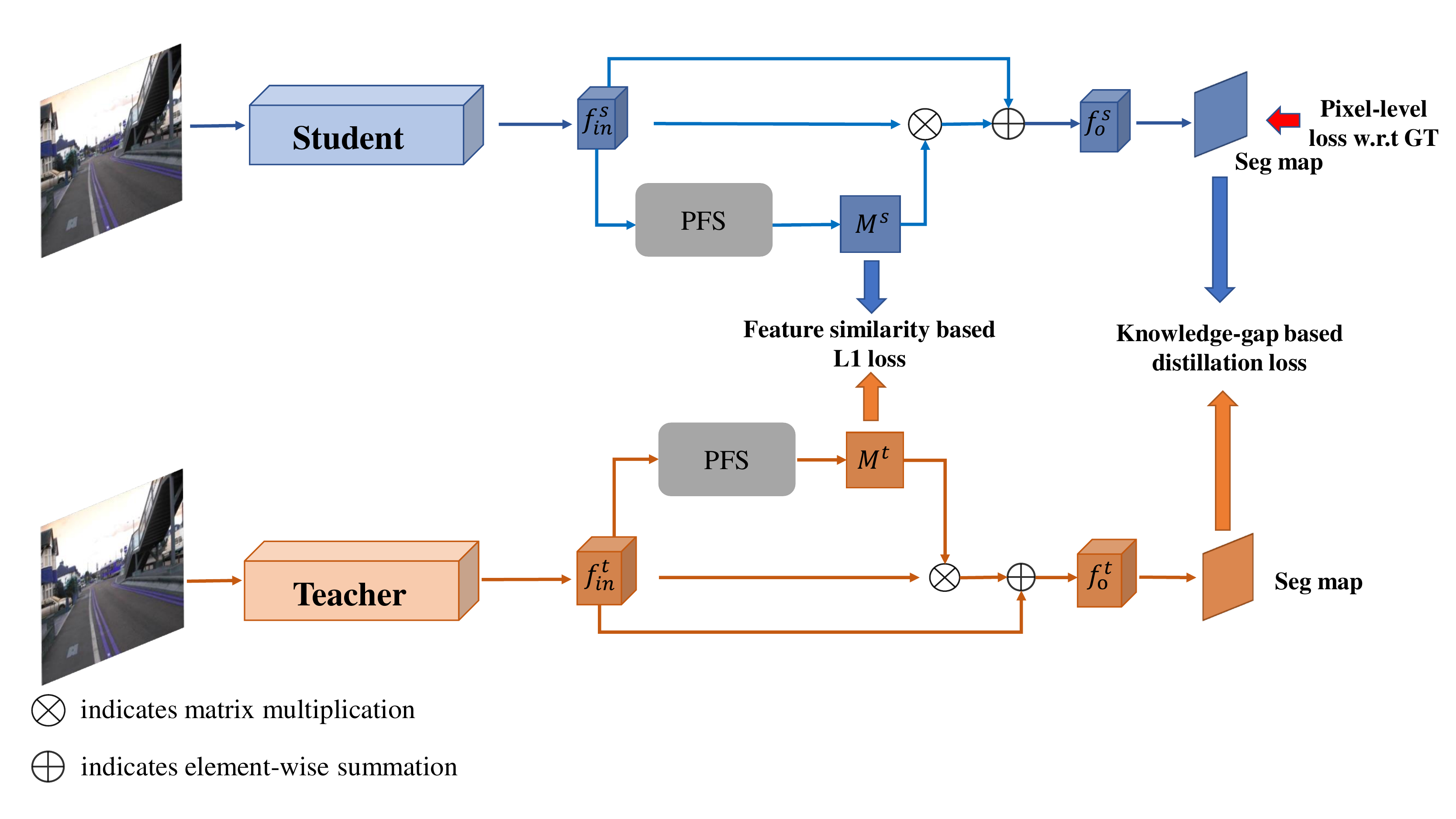}
	\end{center}
	\vspace{-0.8cm}
	\caption{An overview of the proposed knowledge distillation framework. Teacher network with orange color is fixed during the whole distillation process. Student network is colored by blue and optimized by three losses during the distillation process, which are the original segmentation loss, feature similarity based L1 loss and the knowledge gap based distillation loss.}
	\label{fig2}
	\vspace{-0.5cm}
\end{figure*}
Knowledge distillation~\cite{hinton2015distilling,romero2014fitnets,urban2016deep,ashok2017n2n,chen2015net2net,yim2017gift,zagoruyko2016paying,huang2017like,xu2018training} involves a teacher-student strategy where a larger network trained for a given task teaches a smaller network on the same task. \cite{bucilua2006model} trains a compressed model to mimic the output of an ensemble of strong classifiers. \cite{hinton2015distilling} introduces a temperature cross entropy loss in the soft prediction imitation during training. \cite{ba2014deep} proposes to regress the logits (pre-softmax activations) where more information is maintained. \cite{romero2014fitnets} introduces FitNets which extends the soft prediction imitation to the intermediate hidden layers and shows that mimicking in these layers can improve the training of deeper and thinner student network. \cite{chen2015net2net} uses a teacher-student  network system with a function-preserving transform to initialize the parameters of the student network. \cite{ashok2017n2n} compresses a teacher network into an equally capable student network based on reinforcement learning. \cite{yim2017gift} defines a novel information metric for knowledge transfer between the teacher and student network. \cite{zagoruyko2016paying} uses the attention map summarized from the CNN features to achieve knowledge transfer. \cite{huang2017like} explores a new kind of knowledge neuron selectivity to transfer the knowledge from the teacher model to the student model and achieves better performance. 

Beyond the above works that utilize knowledge distillation to improve the performance of student networks for image classification, some works extend knowledge distillation to more complex vision tasks, \emph{e.g.}, object detection~\cite{li2017mimicking,chen2017learning}, face verification and alignment~\cite{wang2017model,luo2016face} and pedestrian detection~\cite{shen2016teacher}. \cite{fastseg} is the only one we find currently dealing with the semantic segmentation task. Specifically, the authors take the probability output and the segmentation boundaries as the transferred knowledge. In contrast, our proposed approach performs pixel-wise distillation, in which each pixel obtains its knowledge of similarities with respect to other ones. 

\section{Proposed method} 
In this section, we present the details of our proposed method. The feature similarity based distillation achieves pixel-wise mimic regarding to the image spatial structure, which is described in Section \ref{self-att}. Knowledge-gap based distillation realizes pixel-wise mimic considering the different necessities of being taught among the pixels, as illustrated in Section \ref{know-gap}. Before starting the knowledge distillation process, we first train a teacher network on the given training set to reach its best performance. Formally, for the task of semantic segmentation, let $X=\{x_{i},y_{i}\}_{i=1}^{N_{train}}$ represents the dataset with $N_{train}$ samples. Pixel-wise multi-class cross entropy loss \cite{long2015fully} regarding to the ground truth label map is employed to update the network. As shown in Figure \ref{fig2}, after the teacher network is well-trained, it is kept fixed to provide knowledge targets to the student. The student network is then driven by the regular segmentation loss as well as the losses generated by the proposed distillation modules.
\subsection{Feature similarity based knowledge distillation} \label{self-att}
Current knowledge distillation methods are mostly for image-level or proposal-level classification (\emph{i.e.}, object detection). In these works, teacher's knowledge is summarized over the whole image or region proposals and then transferred to the student. One naive way of adapting the existing methods to semantic segmentation is to treat the pixel-wise high-dimensional CNN features encoded by the teacher as knowledge transfer media, as done by \cite{romero2014fitnets}. However, only marginal improvements are achieved as shown in our experiments in Section \ref{expe}. To facilitate the distillation process of semantic segmentation, we propose to extract the new knowledge representation encoded by PFS.

\begin{figure}
	\begin{center}
		\includegraphics[width=8cm, height=6cm]{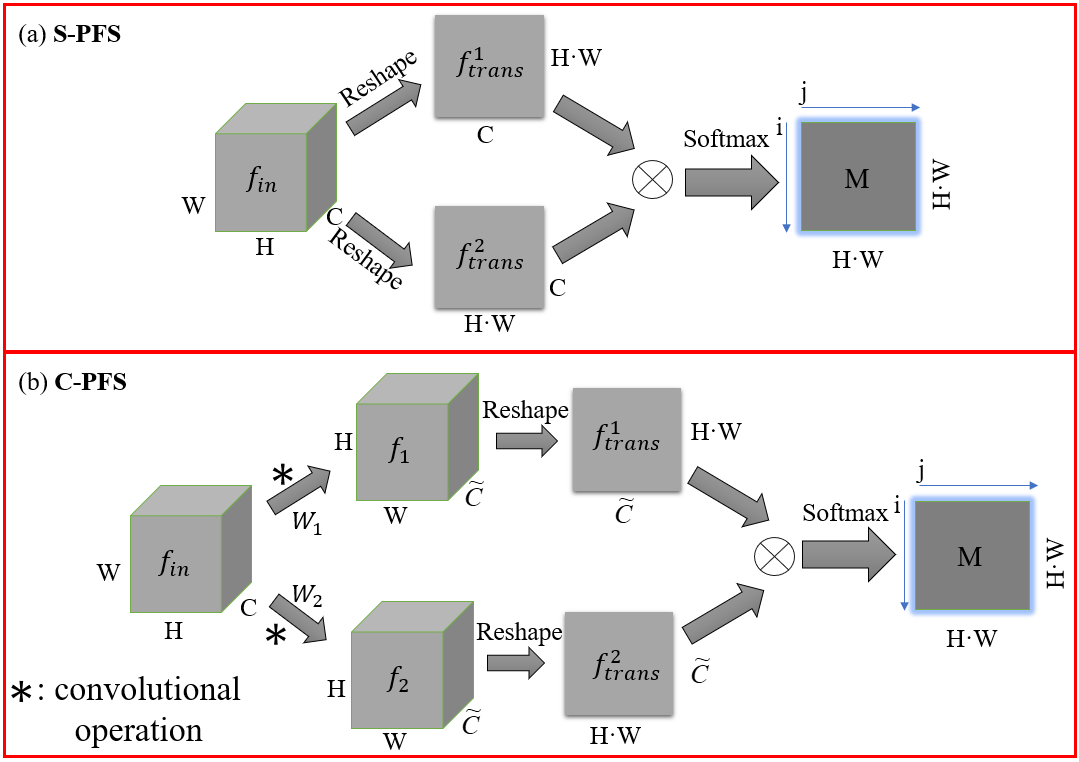}
	\end{center}
	\vspace{-0.3cm}
	\caption{Detailed computing process of the proposed pixel-wise feature similarities. (a) is the simplified version which directly computes PFS on the raw inputs and (b) is the complicated version with two more convolutional transformations on the inputs.}
	\label{fig3}
	\vspace{-0.5cm}
\end{figure}

Take the final convolutional output feature map of the ResNets as an example. We present two methods to extract the pixel-wise feature similarities as shown in Figure \ref{fig3}, the simplified version of PFS (\textbf{S-PFS}) and the complicated version one (\textbf{C-PFS}) which is partially similar to \cite{wang2017non}.
Formally, we denote the convolutional features utilized for the computation of PFS as $f_{in}$ with size of [B, C, H, W]. Here, B, C, H and W represent the batch size, channel number, height and width of the input feature map. For the calculation of \textbf{S-PFS}, raw input $f_{in}$ is directly reshaped into tensors $f_{trans}^{1}$ and $f_{trans}^2$ with size of [B, $H \cdot W$, C] and [B, C, $H\cdot W$], respectively. Then the two reshaped tensors are multiplied with the following softmax operation to generate the final overall \textbf{S-PFS} map with size of [B, $H\cdot W$, $H\cdot W$]. The whole calculation process can be expressed as following equation \eqref{eq1}  
\begin{equation}
\begin{aligned}
&S=f_{trans}^{1}f_{trans}^{2}, \\
&M_{ij} = \frac{exp(S_{ij})}{\sum_{j=1}^{H\cdot W}exp(S_{ij})},
\end{aligned}
\label{eq1}
\end{equation}
in which $S$ is the intermediate \textbf{S-PFS} logits before the softmax operation.
As shown in Figure \ref{fig3}, the final obtained output $M$ represents the overall similarities of any two feature locations. Specifically, we extract the knowledge representation for the $i$-th feature location as PFS$^{i}=(M_{i1}, M_{i2}, ..., M_{ij}, ...)$.
For the calculation of \textbf{C-PFS}, we add two extra convolutional operations to the raw inputs firstly to produce two new features $f_1$ and $f_2$, where $f_1=W_1*f_{in}$ and $f_2=W_2*f_{in}$. Then $f_1$ and $f_2$ are reshaped as above for the following calculation. For the convolutional transformations $W_1$ and $W_2$, we use filters with kernel size of 1 and output channel size of $\widetilde{C} = C/8$. Other subsequent computing processes are the same as \textbf{S-PFS}. Then the loss for the PFS based knowledge distillation can be written as,
\begin{equation}
\begin{aligned}
&L_{pfs}= \frac{1}{N} {\sum_{i}||{PFS}^{i}_{t}-{PFS}^{i}_{s}||_1},
\end{aligned}
\label{eq3}
\end{equation}   
where $N=H\cdot W$ is the total number of pixel locations of the input feature map, and ${PFS}_{t},{PFS}_{s}$ indicate the feature similarities extracted by the teacher and student network, respectively.
After obtaining the feature similarity map $M$, we calculate the final output feature $f_o$ with following equation \eqref{eq2}.
\begin{equation}
\begin{aligned}
&f_o = f_{in} + \gamma(f_{in}M^T),  
\end{aligned}
\label{eq2}
\end{equation}
in which $\gamma$ is a learnable parameter to match the scales of the two features. 

\subsection{Knowledge-gap based knowledge distillation} \label{know-gap}
Beyond the PFS based knowledge distillation part, we also propose another novel knowledge-gap based distillation approach for soft prediction mimicking. We first review the conventional knowledge distillation on soft predictions. Let $t$ and $s$ be the teacher and student network, and their pre-softmax scores are denoted as ${Z_{t}}$ and ${Z_{s}}$. Assuming $T$ to be a temperature parameter used to soften the teacher's scores, the post-softmax predictions of the teacher and student  are $\mathbf{P_{t}} = softmax(\frac{{Z_{t}}}{T})$ and $\mathbf{P_{s}} = softmax({Z_{s}})$, respectively. The student is trained to minimize the sum of the cross-entropy w.r.t.\ ground truth label and the cross-entropy w.r.t.\ the soft predictions of the teacher network, as shown with $\mathcal{H}_{hard}$ and $\mathcal{H}_{soft}$ separately in equation \eqref{eq4}. Here the superscript $i$ in $P_{t}^{i}$ and $P_{s}^{i}$ denotes the prediction on the $i$-th class among all the $c$ classes, and $\mu$ is a hyper-parameter to balance both cross-entropies,
\begin{equation}
\begin{aligned}
&L_{cls} = \mathcal{H}_{hard}(\mathbf{P_{s}},\mathbf{y})+\mu \mathcal{H}_{soft}(\bf{P_{s}}, \bf{P_{t}}), \\
& \mathcal{H}_{soft}= -\sum_{i=1}^{c}\bf{P_{t}^{i}}\log P_{s}^{i}, \\
& \mathcal{H}_{hard}= -\sum_{i=1}^{c}\bf{\mathds{1}_{[i=y]}}\log P_{s}^{i},                             
\end{aligned}
\label{eq4}
\end{equation}
in which $y$ indicates the ground truth label among the $c$ categories.
Directly extending the image-level soft prediction mimicking with equation \eqref{eq4} to a pixel-level one is a straightforward adaptation for the dense semantic segmentation. However, it also ignores the varying necessity of being taught by the teacher for each pixel in the student network, as the knowledge-gaps between both networks on different pixels can be greatly diverse. The predictions on easy pixels of both networks may be similar, while there may exist huge differences between the predictions of both networks on the difficult pixels.  To allow each pixel in the student network to be selectively taught by the teacher, according to the knowledge-gap between both networks, we propose to add a weight reflecting the pixel-level knowledge-gap between both networks in the $\mathcal{H_{\text{soft}}}$ term. Formally, the loss function is shown in equation \eqref{eq5}. Subscript $n$ denotes the $n$-th pixel among all the $N_{im}$ pixels in the image. $w_{n}$ is the pixel-level weight showing the difference between the teacher's post-softmax prediction on the ground truth class of the $n$-th pixel $p_{t,n}^{*}$ and that of the student network  $p_{s,n}^{*}$. When the teacher produces worse post-softmax prediction (\emph{i.e.}, lower post-softmax prediction value) on the ground truth class than the student for a certain pixel, $w_{n}$ is set to 0 to avoid the knowledge transfer from the teacher to student in that pixel. 

\begin{equation}
\begin{aligned}
&L_{cls}=\sum_{n=1}^{N}\mathcal{H}_{hard}^n(\mathbf{P_{s,n}},\mathbf{y_{n}})+\mu w_{n} \mathcal{H}_{soft}^n(\bf{P_{s,n}}, \bf{P_{t,n}}) \\
& \mathcal{H}_{hard}^n= -\sum_{i=1}^{c}\bf{\mathds{1}_{[i=y]}}\log P_{s,n}^{i} \\
&\mathcal{H}_{soft}^n = -\sum_{i=1}^{c}\bf{P_{t,n}^{i}}\log P_{s,n}^{i}\\ 
&w_{n} = \max(0, (p_{t,n}^{*}-p_{s,n}^{*})) 
\end{aligned}
\label{eq5}
\end{equation}
During training, the total loss $L$ to be minimized is the sum of $L_{cls}$ and $\lambda L_{pfs}$ shown in equation \eqref{eq_loss}, 
\begin{equation}
L = L_{cls} + \lambda L_{pfs},
\label{eq_loss}
\end{equation}
in which $\lambda$ controls the contributions of PFS based distillation loss to the optimization.
\begin{table*}
	\hspace{-0.5cm}
	\newcommand{\tabincell}[2]{\begin{tabular}{@{}#1@{}}#2\end{tabular}}
	\caption{Using ResNet18-ResNet101 as the \emph{student-teacher} pair, we show the mIoU results (\%) on Pascal VOC 2012 validation set with the distillation modules of \textbf{S-PFS} and \textbf{C-PFS}, respectively.}
	\label{table1}
	\setlength{\tabcolsep}{1.9pt}
	\renewcommand{\arraystretch}{0.8}
	\scriptsize
	\centering
	\hspace{-0.35cm}
	\begin{tabular}{c|c|c|c|c|c|c|c|c}
		\toprule 
		& \multirow{2}{*}{Vanilla student}  &	\multirow{2}{*}{\tabincell{c}{Post-softmax \\ mimic}\cite{hinton2015distilling}} & \multirow{2}{*}{\tabincell{c}{Hint learning}\cite{romero2014fitnets}} & \multirow{2}{*}{\tabincell{c}{Attention transfer}\cite{zagoruyko2016paying}} &  \multirow{2}{*}{\tabincell{c}{\textbf{PFS distillation}}} & \multirow{2}{*}{\tabincell{c}{\textbf{Knowl.-gap post-}  \\  \textbf{softmax mimic}}} & \multirow{2}{*}{\tabincell{c}{\textbf{Final}  \\  \textbf{result}}} & \multirow{2}{*}{ Vanilla teacher} \\
		&  &  & & & &  & \\
		\midrule
		\multirow{2}{*}{\tabincell{c}{ ResNet18-ResNet101 \\ \textbf{S-PFS}}} & \multirow{2}{*}{67.24} & \multirow{2}{*}{\tabincell{c}{67.76 \\ (+0.52)}} & \multirow{2}{*}{\tabincell{c}{67.90 \\ (+0.66)}}& \multirow{2}{*}{\tabincell{c}{67.92 \\ (+0.68)}}& \multirow{2}{*}{\tabincell{c}{69.59 \\ (\textbf{+2.35})}} & \multirow{2}{*}{\tabincell{c}{68.65 \\ (\textbf{+1.41})}} & \multirow{2}{*}{\tabincell{c}{70.01 \\ (\textbf{+2.77})}} & \multirow{2}{*}{73.41} \\
		&  & & & & &   & \\
		\midrule
		\multirow{2}{*}{\tabincell{c}{ ResNet18-ResNet101 \\ \textbf{C-PFS}}} & \multirow{2}{*}{68.96} &  \multirow{2}{*}{\tabincell{c}{69.58 \\ (+0.62)}} & \multirow{2}{*}{\tabincell{c}{69.78 \\ (+0.82)}}&  \multirow{2}{*}{\tabincell{c}{69.66 \\ (+0.70)}} & \multirow{2}{*}{\tabincell{c}{ 72.01 \\ (\textbf{+3.05})}} & \multirow{2}{*}{\tabincell{c}{70.29 \\ (\textbf{+1.33})}} & \multirow{2}{*}{\tabincell{c}{72.94 \\ (\textbf{+3.98})}} & \multirow{2}{*}{75.82}\\
		&  &  & & & &   & \\
		\bottomrule	
	\end{tabular}
	\vspace{-0.4cm}
\end{table*}
\begin{table*}
	\hspace{-0.5cm}
	\newcommand{\tabincell}[2]{\begin{tabular}{@{}#1@{}}#2\end{tabular}}
	\caption{ Using ResNet34-ResNet101 as the \emph{student-teacher} pair, we show the mIoU results (\%) on Pascal VOC 2012 validation set with the distillation modules of \textbf{S-PFS} and \textbf{C-PFS}, respectively.}
	\label{table2}
	\setlength{\tabcolsep}{1.9pt}
	\renewcommand{\arraystretch}{0.8}
	\small
	\centering
	\hspace{-0.35cm}
	\begin{tabular}{c|c|c|c|c|c}
		\toprule 
		& \multirow{2}{*}{Vanilla student} &  \multirow{2}{*}{\tabincell{c}{\textbf{PFS distillation}}} & \multirow{2}{*}{\tabincell{c}{\textbf{Knowl.-gap post-}  \\  \textbf{softmax mimic}}} & \multirow{2}{*}{\tabincell{c}{\textbf{Final}  \\  \textbf{result}}} & \multirow{2}{*}{ Vanilla teacher} \\
		&  &  & & & \\
		\midrule
		\multirow{2}{*}{\tabincell{c}{ ResNet34-ResNet101 \\ \textbf{S-PFS}}} & \multirow{2}{*}{69.80} & \multirow{2}{*}{\tabincell{c}{ 71.81 \\(\textbf{+2.01})}} & \multirow{2}{*}{\tabincell{c}{70.83 \\ (\textbf{+1.03})}} & \multirow{2}{*}{\tabincell{c}{72.04 \\ (\textbf{+2.24})}} & \multirow{2}{*}{73.41}\\
		&  &  & & & \\
		\midrule
		\multirow{2}{*}{\tabincell{c}{ ResNet34-ResNet101 \\ \textbf{C-PFS}}} & \multirow{2}{*}{71.72} &  \multirow{2}{*}{\tabincell{c}{74.27 \\ (\textbf{+2.55})}} & \multirow{2}{*}{\tabincell{c}{72.87 \\ (\textbf{+1.15})}} & \multirow{2}{*}{\tabincell{c}{74.89 \\ (\textbf{+3.17})}} & \multirow{2}{*}{75.82} \\
		&  & & & & \\
		\bottomrule	
	\end{tabular}
	\vspace{-0.3cm}
\end{table*}
\section{Experiments} \label{expe}
\subsection{Datasets}
The datasets used in the experiments are the challenging ADE20K, Pascal VOC 2012 and Pascal Context benchmarks. ADE20K is a scene parsing benchmark with 150 classes. It contains 20210 training images and 2000 validation images for the evaluation. Pascal VOC 2012 segmentation benchmark contains 10582 training images (including both original and extra labeled ones \cite{hariharan2011semantic}) and 1449 validation images which are used for testing in this paper. Pascal Context dataset provides dense semantic labels for the whole scene, which has 4998 images for training and 5105 for test. Following \cite{fastseg}, we only choose the most frequent 59 classes in the dataset for evaluation. 

\subsection{Implementation details}
In this work, both the teacher and student networks are dilated ResNet-FCN models which are adapted from the conventional ResNet models~\cite{he2016deep} for semantic segmentation. Specifically, the dilated ResNet has two main modifications compared to the ResNet for image classification. First, the strides are changed from 2 to 1 for the convolution filters with kernel size 3 in the first residual block of the 4-th and 5-th groups. Thus, the output size is downsized to 1/8 of the original input image. Second, all the convolution layers in the 4-th and 5-th groups of ResNet are casted into dilated convolution layers with dilation rate being 2 and 4, respectively. The developed PFS extraction module is inserted into the 5-th group of the backbone network and participates into the segmentation as introduced in Figure \ref{fig3}.  

Following previous works \cite{zhao2017pyramid,chen2018deeplab}, we also employ the poly learning rate policy where the initial learning rate is multiplied by $(1-\frac{iter}{total\_iter})^{0.9}$ after each iteration. The base learning rate is set to 0.01 for all the used datasets. Balance weights $\mu$ and $\lambda$ are set to 1 and $10^3$ separately to make the corresponding losses into the same order of magnitude. Stochastic Gradient Descent (SGD) optimizer is used to update the network for 60 epochs, with momentum and weight decay parameters set as 0.9 and 0.0001, respectively. General data augmentation methods are used in network training, such as randomly flipping and random scaling (scales between 0.5 to 1.5) the input images.
Mean IoU is used as the metric to evaluate our algorithm as well as other methods.
\subsection{Baseline methods} 
\label{traditional-kt}
To compare with the traditional knowledge distillation methods, here we choose several excellent distillation methods as our baseline and quantitatively compared with our proposed method.
\vspace{-0.5cm}
\paragraph{Post-softmax mimic} Hinton et.al. \cite{hinton2015distilling} proposed to use the soft probability distribution (as shown by equation \eqref{eq4}) as target for the distillation task of image classification. Here we apply this method to the semantic segmentation task and empirically choose the temperature $T$ of 1 for the Pascal VOC 2012 dataset. 
\vspace{-0.5cm}
\paragraph{Hint learning} Hint learning is proposed in \cite{romero2014fitnets} with using the output of intermediate hidden layer learned by the teacher as target knowledge to improve the training process and final performance of the student network. L2 distance loss is usually used to close the output differences between the student intermediate layers and the corresponding teacher one's. In our experiments, we choose the final residual block of the backbone network as the hint layer. Since the output channel dimensions are different (\emph{e.g.}, 512 for ResNet18/ResNet34 and 2048 for ResNet101), we transform the student network with an extra convolutional layer to match the sizes of both outputs. 
\vspace{-0.5cm}
\paragraph{Attention transfer} Zagoruyko et.al. \cite{zagoruyko2016paying} proposed to use attention as the knowledge to help train the student network. Specifically, they define the attention by conducting operation of summation or maximization on the intermediate CNN features along the channel dimension. In contrast, we represent each pixel's knowledge by its similarities with other ones. Here we use the averaged features generated by the final residual block as the learned knowledge for the distillation process.  
\subsection{Quantitative results}
\subsubsection{Results on Pascal VOC 2012}
Together with above three strong baseline methods, we present the quantitative segmentation results of Pascal VOC 2012 in this section.

We firstly conduct experiments with the $\emph{student-teacher}$ pair of ResNet18-ResNet101 under the distillation modules of \textbf{S-PFS} and \textbf{C-PFS}, respectively. The results are shown in Table \ref{table1}. Generally, our proposed method obtains much better results than all the baseline methods, which certifies the importance of modeling pixel-wise knowledge for the distillation task on semantic segmentation. Specifically, traditional knowledge distillation methods can only improve the mIoU results around 0.7\%, while our proposed method can improve the results of ResNet18 from 67.24\%/68.96\% to 69.59\%(\textbf{+2.35\%})/72.01\%(\textbf{+3.05\%}) separately with the only distillation modules of \textbf{S-PFS}/\textbf{C-PFS}. The knowledge-gap based post-softmax mimic also doubles the improvements obtained by the traditional methods, which also certifies the necessary of paying more attention to the difficult pixels within the image. Combining the two proposed distillation modules, significant improvements of \textbf{2.77\%} and \textbf{3.98\%} are achieved for the student network of ResNet18 with \textbf{S-PFS} and \textbf{C-PFS}, respectively. 
\vspace{-0.3cm}
\paragraph{With different student}
We further show our experimental results with the student network replaced by ResNet34 to check the effectiveness of our proposed methods for different student networks.
As shown in Table \ref{table2}, performance of the ResNet34 network can be improved by \textbf{2.01\%} and \textbf{2.55\%} with the separate \textbf{S-PFS} and \textbf{C-PFS} modules. A final significant improvement of \textbf{2.24\%} and \textbf{3.17\%} are achieved by our proposed method, respectively. 
\vspace{-0.3cm}
\paragraph{With stronger teacher}
Considering the fact that \textbf{C-PFS} based ResNet101 owns better segmentation performance, experiments are conducted in this part to check whether a stronger teacher could help the student further. Here we use \textbf{S-PFS} based ResNet18 and ResNet34 as the student and \textbf{C-PFS} based ResNet101 as the teacher for the experiment. As shown in Table \ref{table3}, after the distillation, results of ResNet18/ResNet34 can be improved from 67.24\%/69.80\% to 70.46\%(\textbf{+3.22\%})/72.87\%(\textbf{+3.07\%}), respectively. With stronger teacher, performance of the student can be further enhanced. 
\begin{table}
	\hspace{-0.5cm}
	\vspace{-0.2cm}
	\newcommand{\tabincell}[2]{\begin{tabular}{@{}#1@{}}#2\end{tabular}}
	\caption{Using \textbf{S-PFS} based student(\emph{i.e.}, ResNet18 and ResNet34) and \textbf{C-PFS} based teacher (ResNet101), we show the mIoU results (\%) on Pascal VOC 2012 validation set.}
	\vspace{0.2cm}
	\label{table3}
	\setlength{\tabcolsep}{1.9pt}
	\renewcommand{\arraystretch}{0.6}
	\small
	\centering
	\hspace{-0.35cm}
	\begin{tabular}{c|c|c|c}
		\toprule 
		& \multirow{2}{*}{Vanilla stu.} &  \multirow{2}{*}{\tabincell{c}{\textbf{PFS distillation}}}  & \multirow{2}{*}{ Vanilla tea.} \\
		&  &  &  \\
		\midrule
		\multirow{2}{*}{\tabincell{c}{ \textbf{S-PFS} ResNet18}} & \multirow{2}{*}{67.24} & \multirow{2}{*}{\tabincell{c}{ 70.46 \\(\textbf{+3.22})}} & \multirow{2}{*}{75.82}\\
		&  &  &  \\
		\midrule
		\multirow{2}{*}{\tabincell{c}{ \textbf{S-PFS} ResNet34}} & \multirow{2}{*}{69.80} &  \multirow{2}{*}{\tabincell{c}{72.87 \\ (\textbf{+3.07})}} & \multirow{2}{*}{75.82} \\
		&  & &  \\
		\bottomrule	
	\end{tabular}
	\vspace{-0.3cm}
\end{table}
\begin{table*}
	\hspace{-0.5cm}
	\vspace{-0.3cm}
	\newcommand{\tabincell}[2]{\begin{tabular}{@{}#1@{}}#2\end{tabular}}
	\caption{Using ResNet18-ResNet101 and ResNet34-ResNet101 as the \emph{student-teacher} pair, we show the mIoU results (\%) on ADE20K validation set with the \textbf{C-PFS} module.}
	\vspace{0.3cm}
	\label{table4}
	\setlength{\tabcolsep}{1.9pt}
	\renewcommand{\arraystretch}{0.8}
	\small
	\centering
	\hspace{-0.3cm}
	\begin{tabular}{c|c|c|c|c|c}
		\toprule 
		& \multirow{2}{*}{Vanilla student}  &  \multirow{2}{*}{\tabincell{c}{\textbf{PFS distillation}}} & \multirow{2}{*}{\tabincell{c}{\textbf{\emph{Knowl.-gap} post-}  \\  \textbf{softmax mimic}}} & \multirow{2}{*}{\tabincell{c}{\textbf{Final}  \\  \textbf{result}}} & \multirow{2}{*}{ Vanilla teacher} \\
		&  &  & & &   \\
		\midrule
		\multirow{2}{*}{\tabincell{c}{ ResNet18-ResNet101}} & \multirow{2}{*}{33.43} &  \multirow{2}{*}{\tabincell{c}{ 36.58 \\ (\textbf{+3.13})}} & \multirow{2}{*}{\tabincell{c}{34.74 \\ (\textbf{+1.31})}} & \multirow{2}{*}{\tabincell{c}{37.42 \\ (\textbf{+3.99})}} & \multirow{2}{*}{40.63}\\
		&  &  & & &  \\
		\midrule
		\multirow{2}{*}{\tabincell{c}{ResNet34-ResNet101}} & \multirow{2}{*}{36.44} & \multirow{2}{*}{\tabincell{c}{39.37 \\ (\textbf{+2.93})}} & \multirow{2}{*}{\tabincell{c}{37.53 \\ (\textbf{+1.09})}} & \multirow{2}{*}{\tabincell{c}{39.89 \\ (\textbf{+3.45})}} & \multirow{2}{*}{40.63} \\
		&  & & & & \\
		\bottomrule	
	\end{tabular}
	\vspace{-0.3cm}
\end{table*}
\begin{figure*}
	\captionsetup[subfigure]{labelformat=empty}
	\centering
	\subfloat[]{\includegraphics[width=3.2cm,height=2.2cm]{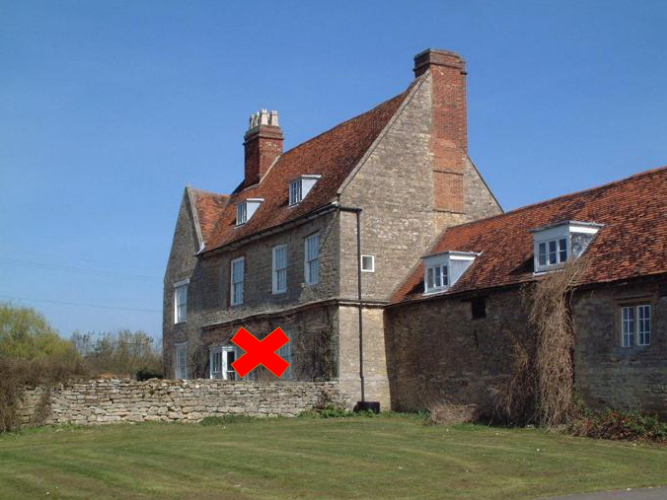}}
	\hspace{0.5cm}
	\subfloat[]{\includegraphics[width=3.2cm,height=2.2cm]{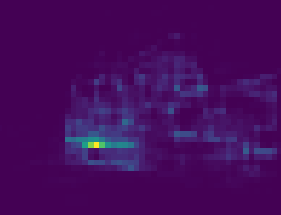}
	}
	\hspace{0.5cm}
	\subfloat[]{\includegraphics[width=3.2cm,height=2.2cm]{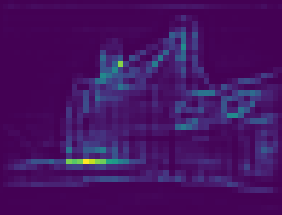}
	}
	\hspace{0.5cm}
	\subfloat[]{\includegraphics[width=3.2cm,height=2.2cm]{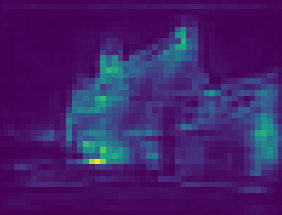}
	}
	\vspace{-0.7cm}
	\subfloat[]{\includegraphics[width=3.2cm,height=2.2cm]{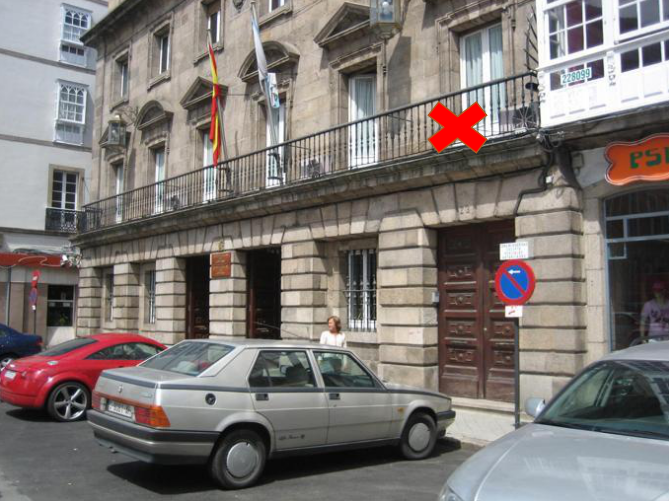}}
	\hspace{0.5cm}
	\subfloat[]{\includegraphics[width=3.2cm,height=2.2cm]{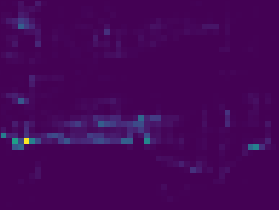}
	}
	\hspace{0.5cm}
	\subfloat[]{\includegraphics[width=3.2cm,height=2.2cm]{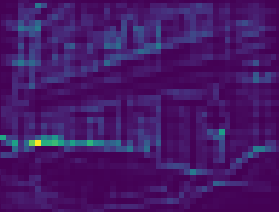}
	}
	\hspace{0.5cm}
	\subfloat[]{\includegraphics[width=3.2cm,height=2.2cm]{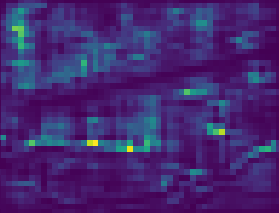}
	}
	\vspace{-0.7cm}
	\subfloat[]{\includegraphics[width=3.2cm,height=2.2cm]{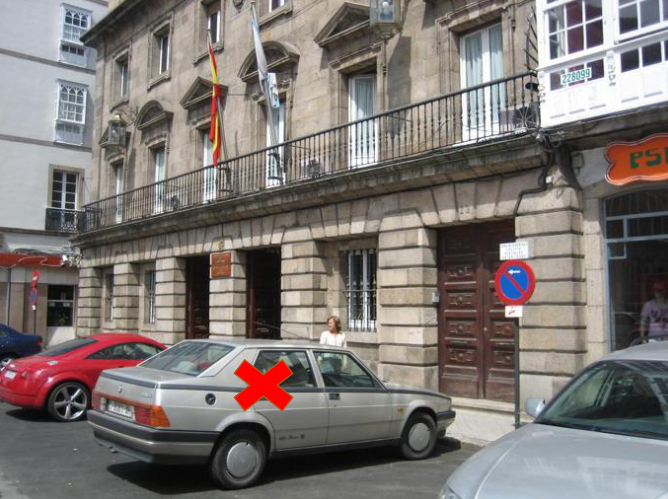}}
	\hspace{0.5cm}
	\subfloat[]{\includegraphics[width=3.2cm,height=2.2cm]{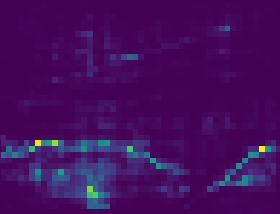}
	}
	\hspace{0.5cm}
	\subfloat[]{\includegraphics[width=3.2cm,height=2.2cm]{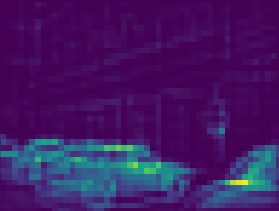}
	}
	\hspace{0.5cm}
	\subfloat[]{\includegraphics[width=3.2cm,height=2.2cm]{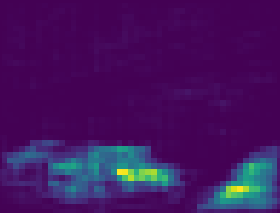}
	}
	\vspace{-0.7cm}
	\subfloat[]{\includegraphics[width=3.2cm,height=2.6cm]{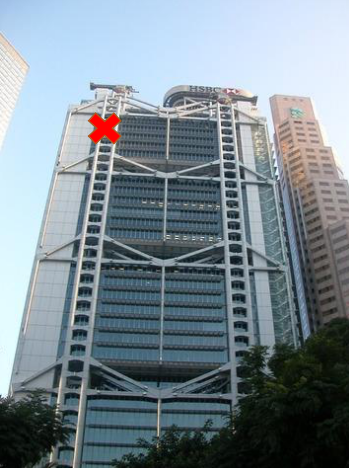}}
	\hspace{0.5cm}
	\subfloat[]{\includegraphics[width=3.2cm,height=2.6cm]{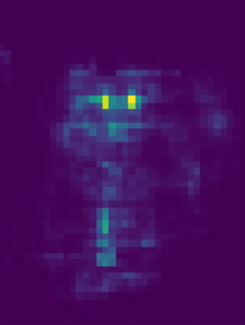}
	}
	\hspace{0.5cm}
	\subfloat[]{\includegraphics[width=3.2cm,height=2.6cm]{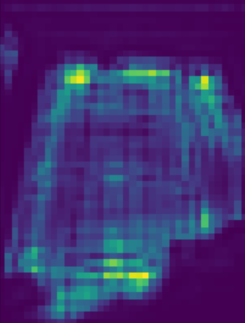}
	}
	\hspace{0.5cm}
	\subfloat[]{\includegraphics[width=3.2cm,height=2.6cm]{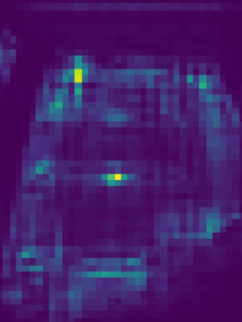}
	}
	\vspace{-0.7cm}
	\subfloat[]{\includegraphics[width=3.2cm,height=2.2cm]{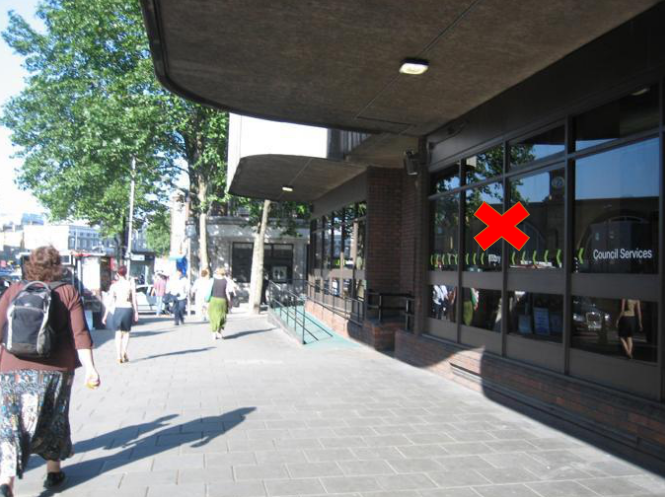}}
	\hspace{0.5cm}
	\subfloat[]{\includegraphics[width=3.2cm,height=2.2cm]{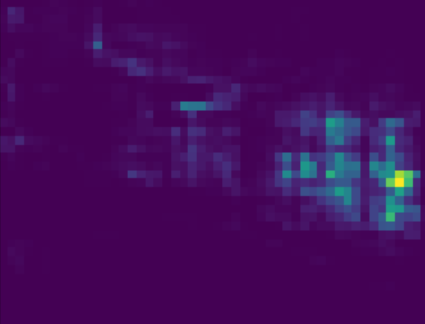}
	}
	\hspace{0.5cm}
	\subfloat[]{\includegraphics[width=3.2cm,height=2.2cm]{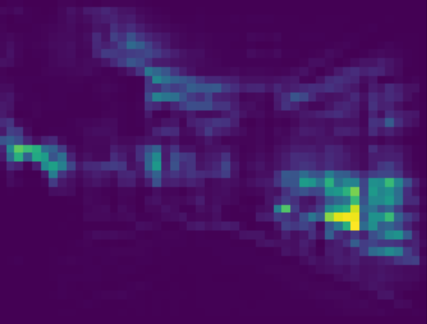}
	}
	\hspace{0.5cm}
	\subfloat[]{\includegraphics[width=3.2cm,height=2.2cm]{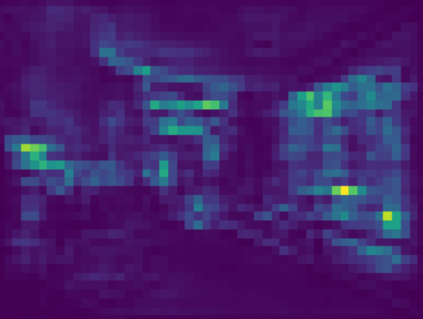}
	}		
	\vspace{-0.5cm}
	\caption{Visualization of the learned PFS at each pixel marked with red cross. From left to right, every column represents the input RGB images, PFS learned by the vanilla student network, distilled student network and the vanilla teacher network, respectively.}
	\label{fig5}
	\vspace{-0.6cm}
\end{figure*}
\subsubsection{Results on ADE20k}
We then list the distillation results of the ADE20k dataset with the \textbf{C-PFS} module in Table \ref{table4}. Consistent improvements are also achieved under the two employed $\emph{student-teacher}$ pairs.
Specifically, the mIoU results are improved from 33.43\% to 37.42\% (\textbf{+3.99\%}) with ResNet18 and 36.44\% to 39.89\% (\textbf{+3.45\%}) with ResNet34. 
Surprisingly, our final result of ResNet34 can even approach its teacher network with only a performance gap of \textbf{0.74\%}. This clearly demonstrates the effectiveness of our proposed method, since the performance gap between the teacher (ResNet101) and the student (ResNet34) is only \textbf{4.19\%}. 
\begin{figure*}
	\captionsetup[subfigure]{labelformat=empty}
	\centering
	\subfloat[]{\includegraphics[width=3.2cm,height=2.2cm]{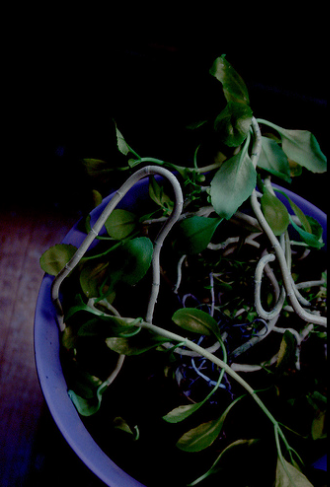}}
	\hspace{0.5cm}
	\subfloat[]{\includegraphics[width=3.2cm,height=2.2cm]{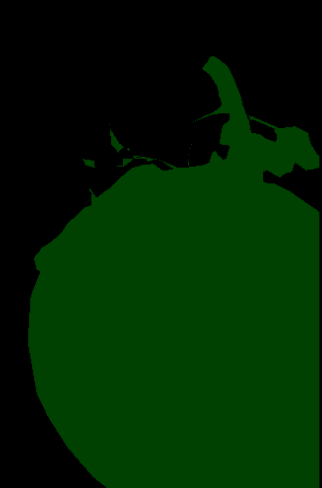}
	}
	\hspace{0.5cm}
	\subfloat[]{\includegraphics[width=3.2cm,height=2.2cm]{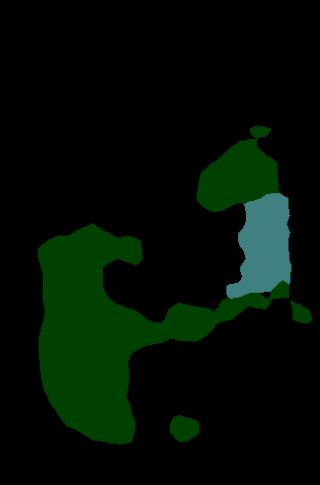}
	}
	\hspace{0.5cm}
	\subfloat[]{\includegraphics[width=3.2cm,height=2.2cm]{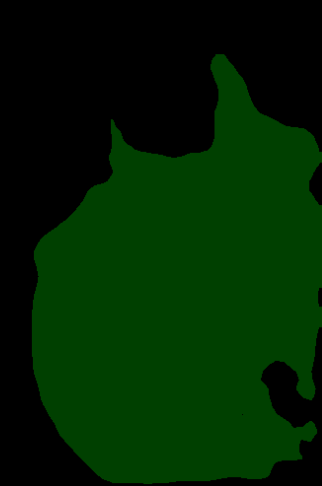}
	}
	\vspace{-0.7cm}
	\subfloat[]{\includegraphics[width=3.2cm,height=2.6cm]{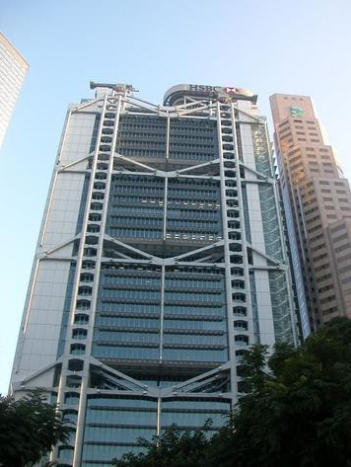}}
	\hspace{0.5cm}
	\subfloat[]{\includegraphics[width=3.2cm,height=2.6cm]{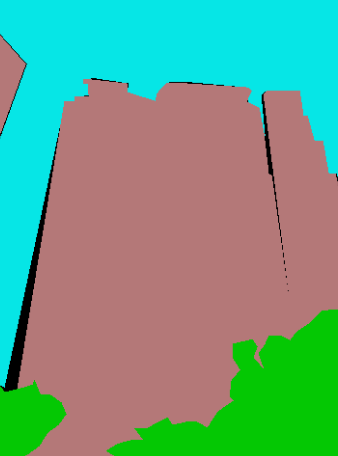}
	}
	\hspace{0.5cm}
	\subfloat[]{\includegraphics[width=3.2cm,height=2.6cm]{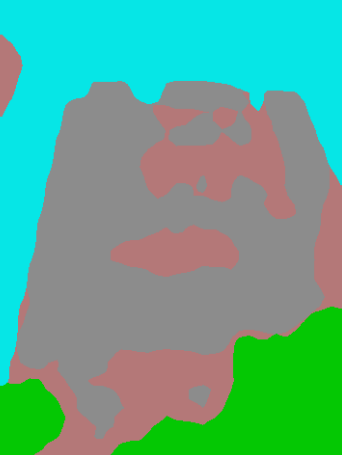}
	}
	\hspace{0.5cm}
	\subfloat[]{\includegraphics[width=3.2cm,height=2.6cm]{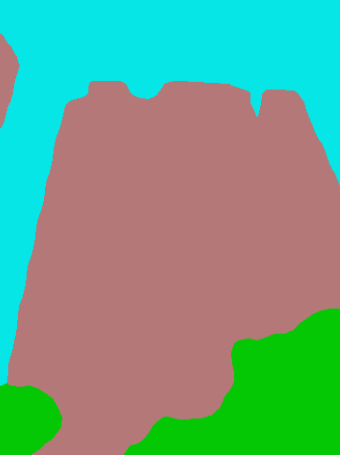}
	}
	\vspace{-0.7cm}
	\subfloat[]{\includegraphics[width=3.2cm,height=2.2cm]{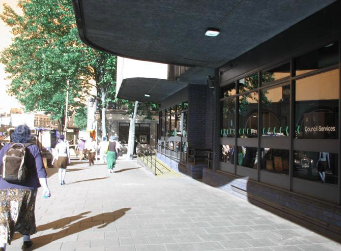}}
	\hspace{0.5cm}
	\subfloat[]{\includegraphics[width=3.2cm,height=2.2cm]{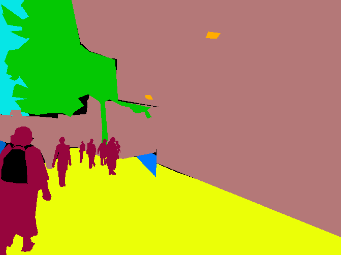}
	}
	\hspace{0.5cm}
	\subfloat[]{\includegraphics[width=3.2cm,height=2.2cm]{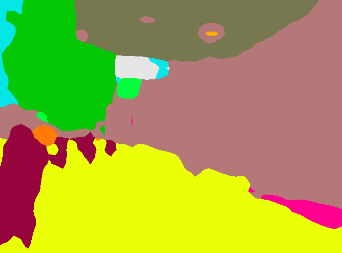}
	}
	\hspace{0.5cm}
	\subfloat[]{\includegraphics[width=3.2cm,height=2.2cm]{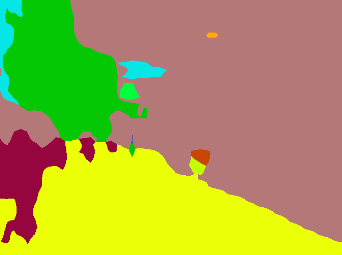}
	}
	\vspace{-0.5cm}
	\caption{Visualization of the segmentation results on Pascal VOC 2012 and ADE20k with our proposed knowledge distillation method. From left to right, we show the input RGB image, ground truth label, segmentation output of the vanilla student network and result of the distilled student network with our proposed method.}
	\label{fig6}
	\vspace{-0.3cm}
\end{figure*}
\begin{table*}
	\hspace{-0.5cm}
	\vspace{-0.3cm}
	\newcommand{\tabincell}[2]{\begin{tabular}{@{}#1@{}}#2\end{tabular}}
	\caption{Comparison with current state-of-the-art mIoU results on the Pascal VOC 2012 and Pascal Context datasets under the \emph{student-teacher} setting of MobileNet1.0-DeepLabV2 and ResNet101-DeepLabV2. We use \textbf{S-PFS} based student and \textbf{C-PFS} based teacher in our training method.}
	\vspace{0.3cm}
	\label{table5}
	\setlength{\tabcolsep}{1.9pt}
	\renewcommand{\arraystretch}{0.5}
	\small
	\centering
	\hspace{-0.35cm}
	\begin{tabular}{cccccccc}
		\toprule 
		\multirow{2}{*}{Datasets}& \multirow{2}{*}{Model}& \multirow{2}{*}{Vanilla student}  &  \multirow{2}{*}{\tabincell{c}{Vanilla teacher}} &  \multirow{2}{*}{\tabincell{c}{Tea.-stu. gap}} & \multirow{2}{*}{\tabincell{c}{Distilled stu.}} & \multirow{2}{*}{\tabincell{c}{Improvements}} &\multirow{2}{*}{Speed(FPS)} \\
		&  &  &  & & & & \\
		\midrule
		\multirow{4}{*}{Pascal Context} & \multirow{2}{*}{\tabincell{c}{\cite{fastseg}}} & \multirow{2}{*}{40.9\%} &  \multirow{2}{*}{\tabincell{c}{48.5\%}} &\multirow{2}{*}{7.6\%}& \multirow{2}{*}{\tabincell{c}{42.8\%}} & \multirow{2}{*}{\tabincell{c}{1.9\%}} & \multirow{2}{*}{\textbf{46.5}}\\
		&  &  & & & & & \\
		~ &\multirow{2}{*}{\tabincell{c}{\textbf{Our method}}} & \multirow{2}{*}{42.1\%} & \multirow{2}{*}{\tabincell{c}{49.8\%}} & \multirow{2}{*}{7.7\%}& \multirow{2}{*}{\tabincell{c}{\textbf{45.2\%}}} & \multirow{2}{*}{\tabincell{c}{\textbf{3.1\%}}} & \multirow{2}{*}{43.8} \\
		&  & & &  & & & \\
		\midrule
		\multirow{4}{*}{Pascal VOC12} & \multirow{2}{*}{\tabincell{c}{\cite{fastseg}}} & \multirow{2}{*}{67.3\%} &  \multirow{2}{*}{\tabincell{c}{75.2\%}} &  \multirow{2}{*}{7.9\%}&
		\multirow{2}{*}{\tabincell{c}{69.6\%}} & \multirow{2}{*}{\tabincell{c}{2.3\%}} & \multirow{2}{*}{\textbf{46.5}}\\
		&  &  & & & & & \\
		~ & \multirow{2}{*}{\tabincell{c}{\textbf{Our method}}} & \multirow{2}{*}{69.1\%} & \multirow{2}{*}{\tabincell{c}{77.3\%}} &  \multirow{2}{*}{8.2\%}&
		\multirow{2}{*}{\tabincell{c}{\textbf{72.7\%}}} & \multirow{2}{*}{\tabincell{c}{\textbf{3.6\%}}} & \multirow{2}{*}{43.8} \\
		&  & & &  & & & \\
		\bottomrule	
	\end{tabular}
	\vspace{-0.3cm}
\end{table*}
\subsection{Qualitative results} 
Since our key idea is to represent the distillation knowledge for each pixel by modeling its similarities with all the other ones, we qualitatively visualize the learned PFS by the vanilla student (\emph{i.e.}, \textbf{S-PFS} based ResNet18), distilled version of the student and the vanilla teacher network (\emph{i.e.}, \textbf{C-PFS} based ResNet101) separately in Figure \ref{fig5}. Different pixels within the image are selected and marked with red bold cross to check their PFS with respect to the whole image contents. Student network given its limited model capacity and representation ability is hard to capture the overall structural information of the image. After conducting the proposed pixel-wise distillation, PFS of the student network can perform similarly with its teacher of including more broaden and long-range image contexts for semantic segmentation. Figure \ref{fig6} shows several example images from the datasets of Pascal VOC 2012 and ADE20K, which include the segmentation results generated by the vanilla student network and the one enhanced by our newly proposed knowledge distillation method. Better segmentation outputs are achieved by the distilled student network after learning the rich feature similarity information from the teacher, which also further certifies the efficacy of our proposed method.        
\subsection{Comparison with state-of-the-arts}
Apart from the knowledge distillation methods introduced in Section \ref{traditional-kt}, \cite{fastseg} is the only one we found to tackle the specific distillation problem of semantic segmentation. In their work, $\emph{student-teacher}$ pair of MobileNet1.0-DeepLabV2 and ResNet101-DeepLabV2 is used to evaluate the segmentation performance. For fair comparison, we implement our method with the same network architectures of \cite{fastseg}. Experiments are conducted on both Pascal VOC 2012 and Pascal Context datasets for complete comparison. In our experiments, we use \textbf{S-PFS} based student and \textbf{C-PFS} based teacher for the distillation and show the comparison results in Table \ref{table5}. Our method achieves consistent better results than current state-of-the-art work. On the Pascal Context dataset, \textbf{S-PFS} based MobileNet1.0 can obtain \textbf{3.1\%} improvement by our proposed method. Under the similar teacher and student performance gap, \cite{fastseg} can only achieve an improvement of 1.9\%. This clearly demonstrates the efficacy of our proposed method. When using the Pascal VOC 2012 dataset, performance of the student network can reach \textbf{72.7\%} with a relative improvement of \textbf{3.6\%}, which both surpasses the results of \cite{fastseg}, \emph{i.e.}, 69.6\% and 2.3\%. Our proposed \textbf{S-PFS} module only takes minor computation time. Compared to \cite{fastseg}, inference speed becomes slightly slower than from 46.5 frames per second (FPS) to 43.8 FPS. However, we can achieve much higher improvements on the mIoU results and still keep the real time segmentation speed.     
\section{Conclusion}
In this paper, we propose a novel pixel-wise feature similarity based knowledge distillation method tailored to the task of semantic segmentation. Our newly designed PFS module endows each pixel with the specific knowledge representation modeled by its similarities with all the other ones, which naturally reflects the inherent structural information of the input image and helps guide the distillation process in an easier way. Furthermore, we propose a novel knowledge-gap based distillation approach on the soft predictions to achieve the weighted knowledge transfer for the pixels, based on the necessity of the student work to be taught by the teacher network. Extensive evaluations on the challenging ADE20K, Pascal VOC 2012 and Pascal Context benchmarks strongly validate the efficacy of our approach in knowledge distillation for semantic segmentation.

\newpage
{\small
	\bibliographystyle{ieee}
	\bibliography{distill}
}
\end{document}